%% file: iclr2027_conference.tex
\documentclass{article} 
\usepackage{iclr2027_conference,times}

\input{math_commands.tex}

\usepackage{hyperref}
\usepackage{url}
\usepackage{booktabs}
\usepackage{wrapfig}

\usepackage{todonotes}

\title{\emph{Which the Eye Fears}: Writing with Read-Blindness Explains Massive Activations in Transformers}

\iclrfinalcopy

\author{Swagatam Mukhopadhyay\\
PsiDagger\\
swag@vislesy.com
\And
Vishal Vivek Saley\\
IIT Delhi\\
vishal.vivek.saley@cse.iitd.ac.in
\And
Vraj Parikh\\
IIT Delhi
\And
Mausam\\
IIT Delhi}

\newcommand{\loss}{\mathcal{L}}

\newcommand{\Mset}{\mathcal{M}}

\DeclareMathOperator{\diag}{diag}

\DeclareMathOperator{\tr}{tr}

\newcommand{\WG}{W_{\text{gate}}}
\newcommand{\WU}{W_{\text{up}}}
\newcommand{\WD}{W_{\text{down}}}

\begin{document}

\maketitle
\fancyhf{}
\renewcommand{\headrulewidth}{0pt}
\renewcommand{\footrulewidth}{0pt}

\input{sections/1_abstract}
\input{sections/2_introduction} 
\input{sections/3_related_works}
\input{sections/4_methodology}
\input{sections/5_read_blindness}
\input{sections/6_amplifier}
\input{sections/7_gradient_analysis}
\input{sections/8_conclusion.tex}
\bibliography{iclr2027_conference}
\bibliographystyle{iclr2027_conference}

\appendix
\input{appendices/defs}
\input{appendices/implementation_details.tex}
\input{appendices/sstar.tex}
\input{appendices/gradient_analysis.tex}
\input{appendices/sensitivity.tex}

\end{document}

%% file: math_commands.tex
\usepackage{amsmath,amsfonts,bm}

\def\eqref#1{equation~\ref{#1}}

\def\1{\bm{1}}

\DeclareMathAlphabet{\mathsfit}{\encodingdefault}{\sfdefault}{m}{sl}
\SetMathAlphabet{\mathsfit}{bold}{\encodingdefault}{\sfdefault}{bx}{n}

\newcommand{\E}{\mathbb{E}}

\newcommand{\R}{\mathbb{R}}



%% file: sections/1_abstract.tex
\begin{abstract}
Massive activation features (MAs) in Transformers are extreme-value residual-stream features that persist across layers despite the model's ability to suppress them. Why do they survive? Our investigation using an operator-level mechanistic analysis of attention and feed-forward (FFN) blocks reveals that these blocks systematically ignore MA coordinates while reading, but not while writing; creating a read-write asymmetry that blocks corrective feedback while allowing continued accumulation. We find that \emph{both} attention and feed-forward layers have this read-blindness, and contribute to the emergence and persistence of MAs.

To validate prior work that hypothesized that FFN's amplification ability is the primary reason for MAs \citep{sun2026massive}, we analyze the model checkpoints during learning. Contrary to our expectation, read-blindness \emph{emerges before} FFN amplification, suggesting that it acts upstream in the MA mechanism. We further contribute gradient analysis to link this behavior to surprising asymmetries in the loss landscape, concluding that the model actively maintains this read-blindness. Finally, we find that removing read-blocking at different locations induces compensatory shifts elsewhere, but MAs still persist.
\end{abstract}

%% file: sections/2_introduction.tex
\section{Introduction}

Massive activation features (MAs) are a small proportion of coordinates in the residual stream of a Transformer whose values exceed the rest of the model features by orders of magnitude \citep{bondarenko2023, sun2024, sun2026massive}. They are central to how the model handles the first token of a sequence \citep{xiao2024}, make low-bit quantization difficult \citep{yu2024}, and shape the geometry of key-value cache compression \citep{ge2024}. The most striking thing about them, however, is how stubbornly they persist. An MA at an early layer typically remains massive through nearly the entire network, even though every later attention head and feed-forward block has, in principle, the linear capacity to subtract it back to a normal range. None of them does. Why?

Several answers to this phenomenon have been proposed, ranging from an attention sink that dominates the BOS token \citep{xiao2024, gu2024attention}, to low-cost storage on delimiter tokens as compression valleys \citep{queipo2025attention}, to more recently, early feed-forward blocks acting as a directional amplifier on a small set of channels \citep{sun2026massive}. Though these attempt to identify the origins of massive activations, mechanistic explanations have been either elusive or partial. For example, none of these explain why no subsequent layer reduces the massive activations.

In this paper, we investigate progression of MAs through the key operators (attention, FFN) in the architecture. We first observe that an operator can fail to correct a residual-stream coordinate because either (1) the coordinate lies in its \emph{output null space}: no choice of input produces a non-zero output for that coordinate; or (2) in its \emph{input null space}; the coordinate is not read by the operator, and it produces the same output irrespective of the coordinate's value.
We find empirically that MAs are \emph{read-blocked} (read-blindness) by attention's $W_V$, $W_Q$, and $W_K$ matrices, and by the feed-forward (FFN) block's gate and up-projections ($W_{up}$), but they are \emph{not write-blocked}: every layer continue to add a non-zero contribution to MAs. We argue that this read-write asymmetry is the mechanism that makes MAs persistent through the network. Our observations adds several nuances and expands far beyond the simple hypothesis by \citet{sun2026massive}, which ascribes FFN amplification as the primary reason for MAs. To investigate this thoroughly, we analyze the model checkpoints during learning and find that read-blindness \emph{emerges before} FFN amplification, suggesting that it acts upstream in the MA mechanism.

We next ask whether the observed asymmetry is actively maintained by training or is merely incidental. At a stable optimum, this question can be probed using local curvature via the Hessian of the loss. We examine the Hessian spectrum of the trained model, along parameters governing read and write pathways. We discover a pronounced asymmetry: directions associated with read operators are \emph{stiff}, exhibiting higher-than-average curvature; while, those associated with write operators are comparatively \emph{sloppy}, with substantially lower curvature. Simply stated, the model appears to actively maintain this read-blindness, and the phenomenon is likely not accidental.

We next attempt two structural interventions to constrain the value projection ($W_V$) so that no null space can develop. Massive activations attenuate, however, the trained model reconfigures and reinforces its read-blindness through other operators such as FFNs. This suggests that MAs are not tied to a single component, but arise from a collusion of mechanisms. Further analysis of FFNs reveals that the previously identified ``amplifier direction'' is generic; it can, in principle, amplify many features; but the down-projection $W_{down}$ selects which of these amplified features are actually written back to the residual stream. It selects several MA features to amplify.

In summary, these results collectively represent a substantial advance in our mechanistic understanding of massive activations in Transformers. Our key result is that massive activations lie in the read-null space of various operators (in attention and FFN blocks), but not in their write-null space, that this behavior is actively maintained by the model, and that blocking read-blindness in one component leads to compensatory behavior in other components.

%% file: sections/3_related_works.tex
\section{Related Work}

\textbf{The Impact and Persistence of Massive Activations:}
Massive activations (MAs); a small proportion of residual stream coordinates that exceed normal representation ranges by orders of magnitude; are a well-documented phenomenon in Transformers \citep{bondarenko2023, sun2024, kaul2024attention, an2025systematic}. They significantly constrain model deployment, acting as primary bottlenecks for low-bit quantization \citep{dettmers2022llmint8, yu2024} and shaping the geometry of KV cache compression \citep{ge2024}. One of the most striking characteristics of MAs is their stubborn persistence across layers. Analyses of their behavior across varied hyperparameters and configurations remain an active area of study \citep{gallego2025hidden, owen2025refined}, yet the question of why subsequent layers fail to normalize these outliers remains a central mystery.

\textbf{Hypotheses on Origins and the FFN Amplifier:} 
Several hypotheses have been proposed to explain the origins and functional utility of MAs. They are frequently tied to how models process the beginning-of-sequence (BOS) token, acting as ``attention sinks'' \citep{xiao2024}, or viewed as low-cost storage functioning as ``compression valleys.'' More recently, a prominent mechanistic explanation by \citet{sun2026massive} posits that early feed-forward (FFN) layers act as directional amplifiers, pushing signals into the massive regime when inputs align with a specific $s^*$ direction. However, our investigation shows that these hypotheses are partial. They primarily address the initiation of MAs but fail to explain their structural persistence; specifically, why subsequent attention heads and FFN blocks do not utilize their linear capacity to subtract these outliers. Furthermore, our empirical findings regarding read-blindness in attention blocks offer a direct counterpoint to the hypothesis that FFN amplifiers are the sole or primary drivers of MAs.

\textbf{Mechanistic Interpretability and Training Dynamics:} 
To mechanistically understand why models stubbornly maintain these outliers, we build upon the spectral and null-space analysis pioneered by \citet{cancedda2024spectral}. We extend this to formally isolate the \emph{input (read)} and \emph{output (write)} null spaces of key operators across Transformer blocks. This formalized lens allows us to uncover the 
read-blindness that explains MA persistence. Finally, to determine if this structural asymmetry is merely incidental or actively maintained, we analyze the trained model through the lens of \emph{stiff} and \emph{sloppy} loss landscapes, measured via Hessian curvature \citep{transtrum1501sloppiness}. This landscape analysis contextualizes our findings that the model actively defends its read-blindness during training, framing MAs as a resilient, distributed mechanism rather than an isolated architectural quirk.

%% file: sections/4_methodology.tex
\section{Preliminaries \& Methodology}\label{sec:methodology}

We investigate the persistence of MAs through the interactions between a Transformer block and the residual stream, which include \emph{reading} from the residual stream through its input projections and \emph{writing} back to it through its output projections. We discover that the persistence of MAs follows from this read--write role asymmetry. As a result, our work creates a framework to systematically evaluate architectural modifications that aim to modify the statistical distribution of feature values, from \emph{massive} to \emph{outlier} and so on, by mechanistically tracking \emph{read} and \emph{write} operations in a Transformer block. 
Formally, given a weight matrix $W$, the \emph{read} component of a linear operation $ \vec{y}= W \vec{x}$ concerns the vector space of $\vec{x}$. \emph{Read blindness} concerns a subspace of $\vec{x}$ not \emph{read} by $W$, i.e., it is the (approximate) right null space of $W$. Similarly, the \emph{write} operation concerns the vector space of $\vec{y}$: \emph{write blindness} implies the presence of an (approximate) left null-space of $W$ preventing it from modifying a subspace of $\vec{y}$. For a general vector space operator, the same concepts hold. The added complexity in Transformers is that the Attention and FFN operators require the residual stream itself as input, before we can study their write operations. We call this the \emph{operator view} (see Sec~\ref{sec:read-blindness}), and general behavior of such operators are reported as average over sampled residual streams. 

\textbf{Transformer architecture.}
We consider a standard Pre-LN Transformer \citep{touvron2023llama} in which each layer applies multi-head attention (MHA) followed by a SwiGLU FFN, with RMSNorm preceding each block. Given residual input $X$, a layer computes $Y=X+\mathrm{MHA}(\widetilde X)$ and $Z=Y+\mathrm{FFN}(\widetilde Y)$, where tildes denote normalized inputs; full
equations are provided in Appendix~\ref{app:notation}. 
This architecture separates the parameters that \emph{read} the residual stream from those that \emph{write} to it. We focus on the MHA read operations coordinated through matrices $W_Q$, $W_K$, and $W_V$ corresponding to the query, key, and value projections because they read the residual stream directly. Likewise, for the the FFN we focus on the reads through the $W_{\text{gate}}$ and $W_{\text{up}}$ matrices. The $W_{\text{down}}$ matrix reads an intermediate vector, namely, the output of SiLU. 
The \emph{read-write} separation allows us to quantitatively isolate and systematically measure the influence of every residual coordinate on the input and output side of every operation, within every block, of every layer of the Transformer. 

\textbf{Massive Activation Coordinates.}
We now formally identify the set $\Mset$ of massively activated coordinates. Let $Z^l_{t,k}(s)$ be the residual coordinate $k$ in the output of layer $l$ and token position $t$ for an input sequence $s$ from a held-out corpus. We define the set of massive coordinates as $\Mset \;=\; \bigl\{\, k : \max_{s,\,l,\,t}\; \tfrac{|Z^l_{t,k}(s)|}{\mathrm{median}_{j}|Z^l_{t,j}(s)|} > R \bigr\}$.
A coordinate belongs to $\Mset$ if it exceeds the threshold at any evaluated sequence, layer, or token position. Unless otherwise specified, we use $R=5$; our results are robust to changes in this threshold (Appendix~\ref{app:sensitivity}). We then compare the read and write behavior of Transformer blocks on coordinates in $\Mset$ and its complement set $\neg\Mset$.

\textbf{Measuring Read-blindness.}
Let $A$ be a given read-side matrix. $A$ is blind to coordinate $k$ when values in coordinate $k$ cause a weak (near-zero) response. For instance, if the standard vector $e_k$ corresponding to coordinate $k$ has $Ae_k$ close to zero, we say that $A$ is blind to coordinate $k$. Based on this intuition, we define the read-blindness (erasure) of a matrix $A$ to coordinate $k$ as follows:
\begin{equation}
\mathrm{era}_k(G) = 1 / (1 + d_m\, G_{kk} / \operatorname{tr}(G)) \;\in\; (0, 1].
\end{equation}
where $G=A^\top A$ is the input-side Gram matrix of $A$. We use Gram matrix instead of the matrix itself as both $G$ and $A$ have the same null space; $G$ being positive semi-definite  makes it a more stable measure for assessing read-blindness. The read-blindness $\mathrm{era}_k(G)$ is close to 1 when $G_{kk}$ is small relative to the trace, and close to 0 when $G_{kk}$ is large. Under uniform treatment across coordinates, the expected score is $0.5$.

To assess whether $\Mset$ has systematically higher erasure scores than a control set $\neg\Mset$, we use a one-sided Mann-Whitney $U$ test on the layer-wise maximum of $\mathrm{era}_k$ for each feature $k$. We use either a size-matched sample or the full complement as specified in each analysis. We define $U$ to count pairwise wins of features in $\Mset$ over features in $\neg\Mset$, assigning half a win to ties, and report the rank-biserial correlation $r_b = 2U/(|\Mset|\,|\neg\Mset|) - 1 \in [-1,+1]$ as an effect-size measure, together with the standardized $z$-score of the $U$ statistic. A value $r_b > 0.95$ with $p \ll 0.05$ indicates that virtually every feature in $\Mset$ has a higher maximum erasure score than virtually every feature in the control set on that Gram matrix.
As a complementary spectral diagnostic, we also measure how strongly each coordinate aligns with the approximate null subspace of $G$. We define this \emph{null occupancy} measure and report its results in Appendix~\ref{app:null-occupancy}.

%% file: sections/5_read_blindness.tex
\section{Read Blindness in Transformers}\label{sec:read-blindness}

We now investigate whether massive coordinates exhibit the read--write asymmetry described above, and whether this behavior persists across model scale and architecture. We analyze three Llama models spanning 135M, 1.28B, and 2.56B parameters, together with a Qwen3 1.7B model. All four models use the Pre-LN residual architecture introduced in Section~\ref{sec:methodology}, with attention and SwiGLU FFN blocks that expose distinct read and write pathways. The Llama models allow us to test whether the phenomenon is stable over an order-of-magnitude change in scale, while Qwen3 tests whether it extends beyond a single model family.

For each model and operator role, Table~\ref{tab:read-write-erasure} reports the median over $k\in\Mset$ of the layer-wise maximum erasure score, with the rank-biserial effect against the full complement $\neg\Mset$ in parentheses. We focus the main-text analysis on this erasure statistic and report the complementary null-occupancy results in Appendix~\ref{app:null-occupancy}. We analyze the learned projections (\emph{Weight View}) and the block inputs and residual updates observed on held-out data (\emph{Operator View}).

\textbf{Weight View.}
On the read side, we apply the input-Gram construction from Section~\ref{sec:methodology} to $W_V$, $W_Q$, $W_K$, and the FFN gate/up projections. The write side instead requires a Gram over residual output coordinates. For attention, let $h$ denote the concatenated head output, so that the residual update is $W_Oh$. We use $G_O^{\mathrm{write}}=W_OW_O^\top$, whose $k$th diagonal entry is $(G_O^{\mathrm{write}})_{kk}=\lVert W_O^\top e_k\rVert_2^2$, the total squared weight through which the attention output can modify residual coordinate $k$. High write-side erasure indicates weak structural access to that coordinate, while low erasure indicates an open write pathway. We apply the same construction to $W_{\mathrm{down}}$; precise definitions of all weight-side Grams are given in Appendix~\ref{app:grams}. Note that Weight View on the write side is an approximation: the actual write operations are dependent on the Attention and FFN operators which are nonlinear functions of the residual. 

\textbf{Operator View.}
As we mentioned above, the write-side weight Grams are an approximation of the actual write operations by the operators of Attention and FFN. The exact results are empirical second moments over a corpus. For attention, $G^{a,oi}=\E[\widetilde X\widetilde X^\top]$ measures mean squared value (we use the shorthand \emph{energy} from here on) of each coordinate at the normalized input, while $G^{a,oo}=\E[(Y-X)(Y-X)^\top]$ measures the energy of the attention residual update at each coordinate. We analogously use $G^{f,oi}=\E[\widetilde Y\widetilde Y^\top]$ and $G^{f,oo}=\E[(Z-Y)(Z-Y)^\top]$ for the FFN in the same way. High erasure on an operator-input Gram means a coordinate carries little magnitude at the input; this differs from erasure on a weight Gram, which means the following block cannot read the coordinate at all. High erasure on an operator-output Gram means the block writes little to that coordinate. The input Grams show whether massive coordinates are present to be read; the output Grams show whether the write pathways identified above are used in practice.

\subsection{Main Results}
\input{tables/detector_unified_table_full.tex}
Massive coordinates are present in the normalized inputs to both blocks. On the operator-input Grams, the $\mathcal{M}$-versus-$\neg\mathcal{M}$ rank-biserial is negative in every model, for both $\widetilde X$ and $\widetilde Y$: massive coordinates are erased less than the non-massive ones, so they carry a larger mean squared value (\emph{energy}) at the input. Their downstream suppression therefore comes from the learned read projections. The signal is present at the input after normalization.

This suppression is strongest and most consistent in $W_V$, $W_Q$, and the FFN gate/up projections. Across all four models, these projections show near-complete separation between massive coordinates and the full complement: $r_b\geq0.99$ for $W_V$ and the FFN projections, and $r_b\geq0.96$ for $W_Q$. The coordinate-wise erasure evidence is weaker for $W_K$, where $r_b$ ranges from $0.11$ to $0.31$. The spectral diagnostic in Appendix~\ref{app:null-occupancy}, however, shows that massive coordinates consistently have greater occupancy in the approximate null subspace of $W_K$, with effects ranging from $r_b=0.86$ for Llama 135M to $r_b=1.00$ for Qwen3 1.7B (Table~\ref{tab:read-write-null-occupancy}). Thus, $W_K$ exhibits strong null-subspace alignment, although its total response to these coordinates is only modestly attenuated.

Both blocks can still write to massive coordinates. For the write
matrices $W_O$ and $W_{\mathrm{down}}$, erasure stays near the uniform
baseline and the $\mathcal{M}$-versus-$\neg\mathcal{M}$ rank-biserial is
negative in every model, so massive coordinates sit inside the write
range of attention and FFN even though the read side suppresses them.
The operator-output Grams agree: the rank-biserial is negative in every
model, so both blocks deposit a larger energy to massive
coordinates than to the non-massive ones on held-out data. Note that because the  second moment is not signed, we cannot tell from it if the deposits coordinate in being additive across layers. We investigate that next.

\paragraph{Signed deposits.}
To investigate signed direction of the deposits acorss layers, we compute $S^b_{\ell,k}=\E_{s,t}[\operatorname{sign}(R^b_{\ell,k})\Delta^b_{\ell,k}]$, using $(R^b,\Delta^b)=(X,Y-X)$ for attention and $(Y,Z-Y)$ for the FFN. Positive values mean that the coordinate's current sign is reinforced whereas negative values mean that they partially cancel. Table~\ref{tab:signed-alignment} reports the median over layers and then over $k\in\Mset$, with the rank-biserial effect against the full complement $\neg\Mset$.

\input{tables/signed_deposit_unified_full_table.tex}

\subsection{Probing on Read-Blindness}\label{sec:interventions}

The previous section showed that several read pathways are blind to massive coordinates. Is blindness in any one of these pathways necessary, or can the model preserve the asymmetry when that pathway is forced open? We test this by preventing either the attention value projection or the FFN input projections from becoming read-blind during training. We then ask whether massive activations disappear and, if not, where the read-blindness reconfigures.

\paragraph{Forcing a read pathway open.}
We train three variants of the Llama 1.28B model using the same hyperparameters as the base model. In the \emph{$W_V$ Frozen} variant, the combined $W_V$ is initialized as a random orthogonal matrix and held fixed throughout training. It therefore has no null space, but also no freedom to learn. The \emph{$W_V$ Reparam} variant instead uses a Cayley parametrization, which allows $W_V$ to learn during training while keeping it orthogonal. In the third variant, \emph{FFN Frozen}, we initialize $W_{\mathrm{gate}}$ and $W_{\mathrm{up}}$ orthogonally and freeze them. We do not use a Cayley parametrization for these large rectangular matrices because of its computational cost and large number of strategies available in such a construction with nothing to motivate a particular choice. Relative to the base-model perplexity of $14.21$, perplexity changes to $14.67$ ($+0.46$) for $W_V$ Frozen, $14.24$ ($+0.03$) for $W_V$ Reparam, and $13.67$ ($-0.54$) for FFN Frozen. Table~\ref{tab:read-write-erasure-intervention} reports the erasure results; the complementary null-occupancy results are reported in Table~\ref{tab:read-write-null-occupancy-intervention} in the appendix.

\paragraph{Read-blindness re-configures elsewhere.}
Both $W_V$ interventions work as intended. The erasure and null-occupancy scores of $W_V$ are $0.50$, with no separation between massive and control coordinates. In the Frozen variant, number of massive coordinates is reduced to 7, and the maximum activation magnitude falls to $2.08\!\times\!10^3$. In the Reparam variant, the number of massive coordinates is 10, and the maximum magnitude is $2.12\!\times\!10^3$, slightly lower compared to the base model. Nevertheless, massive activations remain.
With $W_V$ forced open, read-blindness remains in the other read pathways. In both variants, $W_Q$ and the FFN gate/up projections almost completely separate massive coordinates from their controls, with $r_b\geq0.99$ under both diagnostics. $W_K$ also remains strongly aligned with its approximate null subspace, with null-occupancy effects of $r_b=0.98$ and $0.99$. Thus, removing read-blindness from $W_V$ re-distributes its role across the remaining read operators rather than eliminating it from the model.

The FFN intervention shows the same pattern in the opposite direction. Freezing $W_{\mathrm{gate}}$ and $W_{\mathrm{up}}$ brings both of their blindness scores back to the uniform baseline of approximately $0.50$. Read-blindness remains strong in attention. The effects are $r_b=0.98$ for $W_V$ and $r_b=0.99$ for $W_Q$, while $W_K$ has a null-occupancy effect of $r_b=0.97$. The model still develops 11 massive coordinates, although their maximum magnitude falls to $681$. Opening the FFN read pathway therefore weakens the largest activation, but does not remove massive activations. Because these FFN projections are frozen, part of this reduction may also come from their loss of learning capacity (reduced number of trainable parameters).

The re-distribution is confined to the read side. Across all three variants, $W_O$, $W_{\mathrm{down}}$, and the realized attention and FFN updates show no preferential suppression of massive coordinates. The write pathways remain open and continue to deposit energy into these coordinates.

These interventions demonstrate that no single read projection is responsible for the read--write asymmetry. When one pathway is forced open, read-blindness is re-distributed across the pathways that remain available, while the write side stays open. MAs are therefore maintained and reconfigured in a collaboration between Attention and FFN operators. This result was a surprise to us. 

%% file: tables/detector_unified_table_full.tex
\begin{table}[t]
\centering
\caption{Erasure results. Cells report the median ${\mathrm{era}}_{\mathcal{M}}$ with rank-biserial $r_b$ against the full complement $\neg\mathcal{M}$ in parentheses. Bold cells have $r_b>0.95$; stars denote one-sided Mann--Whitney tests of ${\mathrm{era}}_{\mathcal{M}}>{\mathrm{era}}_{\neg\mathcal{M}}$ with $p<0.05$.}
\label{tab:read-write-erasure}
\small
\setlength{\tabcolsep}{4pt}
\begin{tabular}{rcccc}
\toprule
 & Llama 135M & Llama 1.28B & Llama 2.56B & Qwen3 1.7B \\
\midrule
Massive coordinates $|\mathcal{M}|$ & 10 & 10 & 5 & 7 \\
Maximum activation magnitude & 280 & $2.27\!\times\!10^{3}$ & $3.92\!\times\!10^{3}$ & $2.23\!\times\!10^{3}$ \\
\midrule
\multicolumn{1}{l}{\emph{Input availability:}} &  &  &  &  \\
\quad Attention normalized input, $\tilde{X}$ & 0.58 (-0.60) & 0.68 (-0.73) & 0.55 (-0.64) & 0.58 (-0.37) \\
\quad FFN normalized input, $\tilde{Y}$ & 0.56 (-0.65) & 0.56 (-0.13) & 0.50 (-0.63) & 0.54 (-0.48) \\
\midrule
\multicolumn{1}{l}{\emph{Read side weight view:}} &  &  &  &  \\
\quad Value projection $W_V$ & \textbf{0.70 (+0.99)}\textsuperscript{*} & \textbf{0.75 (+0.99)}\textsuperscript{*} & \textbf{0.78 (+1.00)}\textsuperscript{*} & \textbf{0.76 (+1.00)}\textsuperscript{*} \\
\quad Query projection $W_Q$ & \textbf{0.63 (+0.99)}\textsuperscript{*} & \textbf{0.65 (+0.99)}\textsuperscript{*} & \textbf{0.64 (+0.99)}\textsuperscript{*} & \textbf{0.66 (+0.96)}\textsuperscript{*} \\
\quad Key projection $W_K$ & 0.56 (+0.29) & 0.55 (+0.31)\textsuperscript{*} & 0.53 (+0.11) & 0.59 (+0.25) \\
\quad FFN gate/up $W_{\mathrm{gate}},W_{\mathrm{up}}$ & \textbf{0.66 (+0.99)}\textsuperscript{*} & \textbf{0.63 (+1.00)}\textsuperscript{*} & \textbf{0.59 (+0.99)}\textsuperscript{*} & \textbf{0.62 (+0.99)}\textsuperscript{*} \\
\midrule
\multicolumn{1}{l}{\emph{Write side weight view:}} &  &  &  &  \\
\quad Attention output $W_O$ & 0.51 (-0.82) & 0.51 (-0.61) & 0.49 (-0.59) & 0.51 (-0.15) \\
\quad FFN down projection $W_{\mathrm{down}}$ & 0.49 (-0.86) & 0.49 (-0.64) & 0.49 (-0.46) & 0.49 (-0.88) \\
\midrule
\multicolumn{1}{l}{\emph{Write side operator view:}} &  &  &  &  \\
\quad Attention update $\Delta^{\mathrm{attn}}$ & 0.61 (-0.87) & 0.85 (-0.44) & 0.62 (-0.64) & 0.64 (-0.36) \\
\quad FFN residual update $\Delta^{\mathrm{ffn}}$ & 0.81 (-0.29) & 0.69 (-0.77) & 0.59 (-0.59) & 0.56 (-0.60) \\
\bottomrule
\end{tabular}
\end{table}

%% file: tables/signed_deposit_unified_full_table.tex
\begin{table}[t]
\centering
\caption{Signed-alignment results. For each coordinate, $S$ is median-aggregated across layers. Cells report the median $S_{\mathcal{M}}$ with rank-biserial $r_b$ comparing $\mathcal{M}$ and  $\neg\mathcal{M}$ in parentheses.}\label{tab:signed-alignment}
\small
\setlength{\tabcolsep}{4pt}
\begin{tabular}{lcccc}
\toprule
Signed alignment & Llama 135M & Llama 1.28B & Llama 2.56B & Qwen3 1.7B \\
\midrule
Attention, $X_k$ with $\Delta_k^{\mathrm{attn}}$ & +0.03 (+0.45) & +0.09 (+0.29) & +0.02 (-0.27) & +0.06 (+0.16) \\
FFN, $Y_k$ with $\Delta_k^{\mathrm{ffn}}$ & +0.05 (+0.57) & +0.25 (+0.59) & -0.02 (-0.20) & +0.10 (+0.14) \\
\bottomrule
\end{tabular}
\end{table}

%% file: sections/6_amplifier.tex
\section{Sun's amplifier and the geometry of FFN read-blindness}\label{sec:sstar}

We now investigate the relationship of our observations so far to previous work on FFN  in \citet{sun2026massive}. They blame the origins of massive activations exclusively on a directional FFN amplifier. They formulate FFN's write to coordinate $k$ as a quadratic form in the normalized residual $\widetilde h$.
\begin{equation*}
  y_k(\widetilde h)\approx \widetilde h^\top S_k\,\widetilde h,\qquad
  S_k=\tfrac12(U_k+U_k^\top),\qquad
  U_k=\sum_i \WD[k,i]\,\WG[i,:]^\top\WU[i,:].
\end{equation*}
\citet{sun2026massive} show that under approximation (Appendix~\ref{app:sstar_amplifier}), $\mathrm{FFN}(\widetilde h)_k\approx\lambda_\star^{(k)}(s_\star^{(k)\top}\widetilde h)^2$, where $s_\star^{(k)}$ is the unit input direction (eigen-vector of $S_k$) that drives output coordinate $k$ and $\lambda_\star^{(k)}$ is its signed gain (corresponding eigen-value). Note that since $s_\star^{(k)}$ need not align with the residual coordinate direction $e_k$, the FFN can be \emph{blind} to the value stored in $k$ while still reading $s_\star^{(k)}$ and writing a large update to $k$ proportional to the signed-gain $\lambda_\star^{(k)}$ . This will become important in our analysis.

\paragraph{Amplifier specificity lies in gain and spectral concentration}
Each output coordinate $k$ has an amplifier direction $s_\star^{(k)}$ and a corresponding gain $\lambda_\star^{(k)}$. We ask which of these is distinctive for massive coordinates: do they share a special amplifier direction, or do they instead have unusually strong amplification along their respective directions? We test direction using the within-set absolute cosine, which measures how strongly the amplifier directions for different coordinates align. We test amplifier strength using $\|U_k\|_F$, the overall scale of the quadratic construction, and $|\lambda_\star^{(k)}|$, the gain along $s_\star^{(k)}$. Finally, $|\lambda_\star^{(k)}|/\|S_k\|_F$ measures how strongly the quadratic form is concentrated in that dominant direction. We report an additional subspace-alignment diagnostic in Appendix~\ref{app:sstar_amplifier}.

Table~\ref{tab:sstar-diag} shows that direction is not the main distinction. The within-set absolute cosine is only moderately higher for $\Mset$ than for the size-matched non-massive set $c\Mset\subset\neg\Mset$ ($0.53$ versus $0.46$), so massive coordinates do not share a clearly distinctive amplifier direction. Gain provides a much sharper distinction: $\|U_k\|_F$ and $|\lambda_\star^{(k)}|$ have rank-biserial effects of $0.78$ and $0.98$, respectively. The strongest discriminator is the concentration $|\lambda_\star^{(k)}|/\|S_k\|_F$, which separates the two groups perfectly in this sample ($r_b=1.00$). 

\begin{wraptable}{r}{0.58\textwidth}
  \vspace{-1.7\baselineskip}
  \caption{$s_\star$ amplifier diagnostics for Llama 1.28 Base. Per-coordinate entries are medians over $\mathcal{M}$ and a deterministic, size-matched $\neg\mathcal{M}$; the final row is the layer-mean of the within-set median pairwise absolute cosine. $r_b > 0$ means larger values on $\mathcal{M}$.}\label{tab:sstar-diag}
  \centering
  \small
  \begin{tabular}{lccc}
    \toprule
    Diagnostic & value $\mathcal{M}$ & $r_b$ \\
    \midrule
    $\lVert U_k\rVert_F$ & 7.74 & +0.78 \\
    $\lvert\lambda_\star^{(k)}\rvert$ & 2.19 & +0.98 \\
    $\lvert\lambda_\star^{(k)}\rvert/\lVert S_k\rVert_F$ & 0.39 & +1.00 \\
    Within-set $\lvert\cos(s_\star)\rvert$ & 0.53/0.46 & --- \\
    \bottomrule
  \end{tabular}
  \vspace{-0.8\baselineskip}
\end{wraptable}

We observe that massive coordinates are strongly distinguished by having higher quadratic gain concentrated in a dominant direction, indicating a shift toward approximately rank-one behavior of the amplifier. However, the specific direction in the FFN's internal space is coordinate-specific and is not shared among the massive coordinates. Intuitively, this lack of collusion is perhaps unsurprising because the FFN has a separate quadratic form $S_k$ for each output coordinate. What is surprising is the collapse of these forms toward a single direction with high gain. What occurs first: read-blindness or this collapse? We answer that next.

\paragraph{Read-blindness is detected before amplifier specialization.} We track FFN read-blindness and amplifier gain through training. For each checkpoint and coordinate, we take the maximum across layers of the FFN erasure $\mathrm{era}_k$ derived from FFN's read-side matrices and the amplifier gain $|\lambda_\star^{(k)}|$. We assign $\Mset$ and the non-massive control set using the final checkpoint, and then trace these fixed sets backward through training. This lets us ask which FFN signature first distinguishes the coordinates that eventually become massive. We concede that this analysis tests temporal ordering, not causal dependence.

Figure~\ref{fig:temporal}(a--b) summarizes the comparison. We define the separation time $t_{\rm sep}$ as the first checkpoint at which the rank-biserial effect between eventual massive and control coordinates reaches $r_b\geq0.95$. FFN read-blindness separates the groups at $2.3$k steps (Figure~\ref{fig:temporal}(a)), whereas amplifier gain separates them later, at $4.5$k steps (Figure~\ref{fig:temporal}(b)). Thus, FFN read-blindness is already detectable before the amplifier specializes. This temporal precedence does not establish that read-blindness causes amplifier specialization; it shows that read-blindness is the earlier structural signature of the coordinates that eventually become massive. Together with Section~\ref{sec:read-blindness}, the results support complementary roles: the amplifier explains how selected coordinates receive exceptional writes, while FFN read-blindness explains why subsequent FFN blocks do not condition corrective updates on the massive values already present.

\input{figs/temporal.tex}

\paragraph{$\WD$ concentration and the read-blindness leak.}
The FFN amplifier poses a puzzle. The FFN is read-blind to massive coordinates, so how does it nevertheless produce large writes to them? The blindness is not total. Attention erases massive coordinates more strongly than the FFN: in the 1.28B model, $W_V$ erasure is $0.75$, compared with $0.63$ for $\WG,\WU$ (Table~\ref{tab:read-write-erasure}); the same gap appears through training in Figure~\ref{fig:emergence_full}. This difference is plausible since, unlike attention, the FFN acts independently at each token. The weaker FFN erasure therefore leaves a partial read pathway through which some FFN intermediate channels can still respond to massive residual coordinates.

Under the quadratic approximation, the FFN writes strongly to residual coordinate $k$ when the input $\widetilde h$ aligns with $s_\star^{(k)}$, the top eigenvector of $S_k$. From the sum defining $U_k$, this direction is determined by the read directions $\WG[i,:]$ and $\WU[i,:]$ of the intermediate channels $i$ weighted by $\WD[k,i]$. For the partial read pathway to produce a strong amplifier, two conditions are needed. First, some intermediate channels that contribute to output coordinate $k$ must retain read access to the massive residual coordinates. Second, the write row $\WD[k,:]$ must concentrate on those channels. Then $S_k$ is dominated by a few read directions and can develop a high-gain dominant mode; if $\WD[k,:]$ is spread across many unrelated units, their contributions are diluted.

The hypothesis predicts that FFN gain on coordinate $k$ grows with the
concentration of $\WD[k,:]$. We measure concentration using the inverse participation ratio (IPR) metric defined below.

\begin{equation*}
\mathrm{IPR}\bigl(\WD[k,:]\bigr)=
   \frac{\sum_i \WD[k,i]^4}{\bigl(\sum_i \WD[k,i]^2\bigr)^2}
   \in\bigl[\,1/d_{\rm ffn},\,1\,\bigr]
\end{equation*}

A higher IPR means that the write row for residual coordinate $k$ is concentrated on fewer FFN intermediate units.
Figure~\ref{fig:temporal}(c) reports the result. Within $\Mset$, residual coordinates with more concentrated $\WD$ rows have larger amplifier norm $\|U_k\|_F$, consistent with concentration turning the partial FFN read pathway into a strong amplifier. Why does $\WD[k,:]$ concentrate? We suspect a token-independent coordinate is served by a few dedicated FFN intermediate channels, with weight decay and small-initialization gradient descent favoring the low-norm row that uses only those channels.

%% file: figs/temporal.tex
\begin{figure*}[t]
  \centering

  \begin{minipage}[t]{0.285\textwidth}
    \vspace{0pt}
    \centering
    \includegraphics[width=\linewidth]{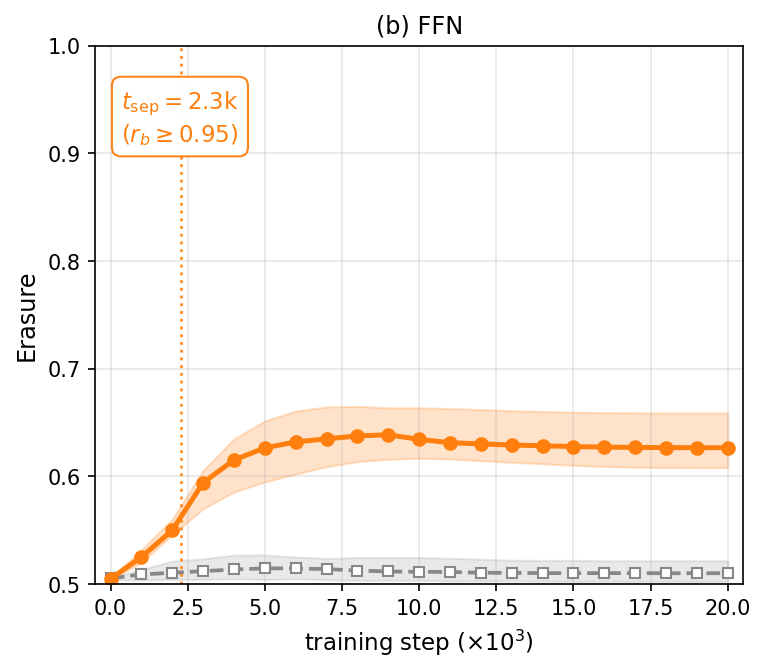}
    \par\smallskip
    \textbf{(a)} FFN read-blindness
  \end{minipage}\hfill
  \begin{minipage}[t]{0.285\textwidth}
    \vspace{0pt}
    \centering
    \includegraphics[width=\linewidth]{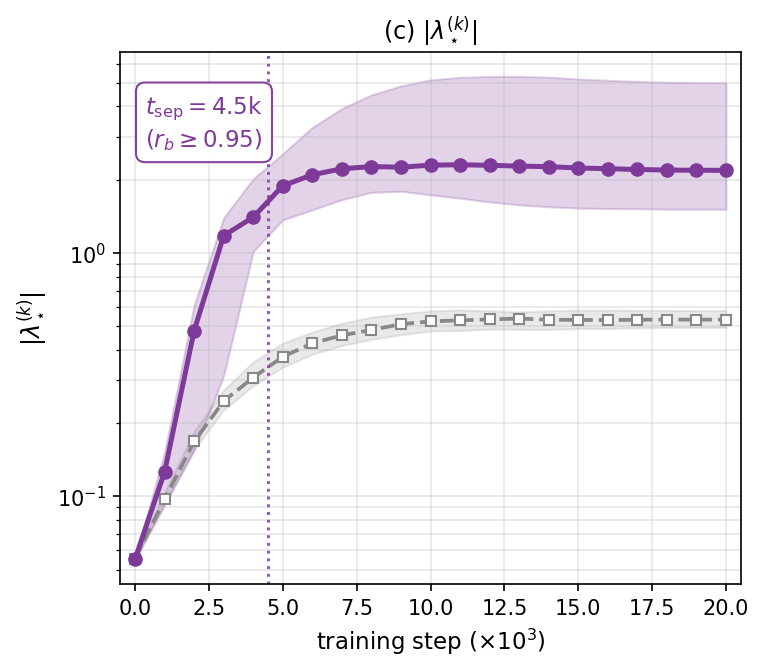}
    \par\smallskip
    \textbf{(b)} Amplifier gain
  \end{minipage}\hfill
  \begin{minipage}[t]{0.39\textwidth}
    \vspace{0pt}
    \centering
    \includegraphics[width=\linewidth]{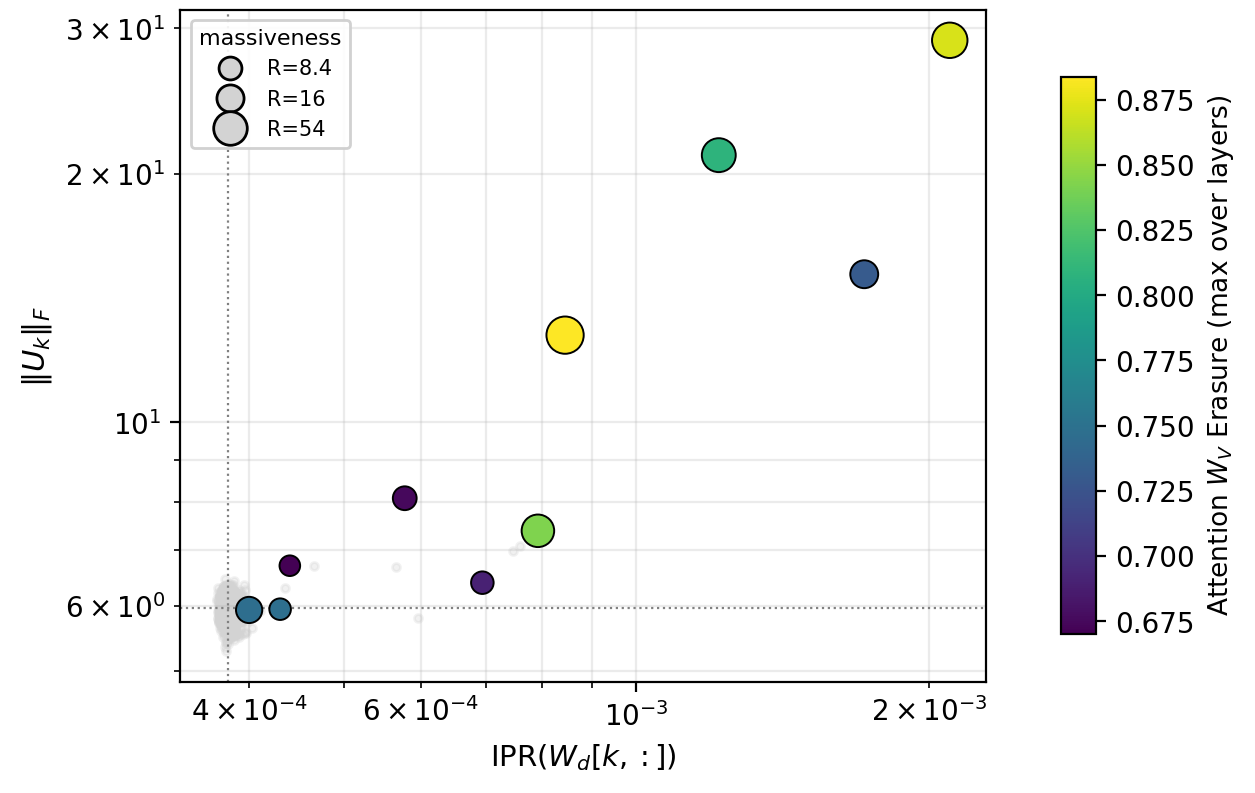}
    \par\smallskip
    \textbf{(c)} Write-row concentration vs. amplifier norm
  \end{minipage}

  \caption{FFN read-blindness precedes amplifier specialization. We assign
  massive and control coordinates at the final checkpoint and trace these
  fixed sets through 21 checkpoints from $0$ to $20$k training steps.
  \textbf{(a)} Maximum-over-layer FFN erasure for eventual massive
  coordinates (orange) and non-massive controls (gray).
  \textbf{(b)} Maximum-over-layer leading amplifier gain for the same sets.
  Lines and shaded regions show medians and interquartile ranges; vertical
  dotted lines mark $t_{\rm sep}$, the first checkpoint at which the
  rank-biserial effect reaches $r_b\geq0.95$. FFN erasure separates at
  $2.3$k steps, before amplifier gain at $4.5$k steps.
  \textbf{(c)} Final-checkpoint amplifier norm $\|U_k\|_F$ versus the inverse
  participation ratio of the corresponding FFN write row $\WD[k,:]$.
  Colored points are massive coordinates, with size denoting massiveness and
  color denoting attention-input erasure; gray points are non-massive
  coordinates.}
  \label{fig:temporal}
\end{figure*}

%% file: sections/7_gradient_analysis.tex
\section{Training actively reinforces massive activations}
\label{sec:gradient-analysis}

The intervention experiments show that the model can reorganize its read
pathways to preserve read-blindness. We now ask whether this configuration is also visible in the local loss geometry, and whether the next optimizer step would preserve or reduce the resulting massive activations.

\textbf{The loss geometry follows the read--write asymmetry.} We borrow the language of \emph{stiff} and \emph{sloppy} directions from
\citet{transtrum1501sloppiness}. A direction is stiff when moving away from trained parameters increases the loss, and sloppy when the parameters can move without incurring much penalty. We measure this sensitivity using the directional Hessian curvature $v^\top H v$. For each parameter slice associated with a residual coordinate, we sample multiple random unit directions $v$ and average their curvature. Positive curvature indicates a locally stable stiff direction, curvature near zero suggests a flat direction, and negative curvature indicates that the parameters do not lie in a locally convex basin along that direction. Appendix~\ref{sec:def:curvature} gives the full definition and sampling procedure.

\begin{wrapfigure}{r}{0.5\textwidth}
    \vspace{-0.75\baselineskip}
    \centering
    \includegraphics[width=\linewidth]{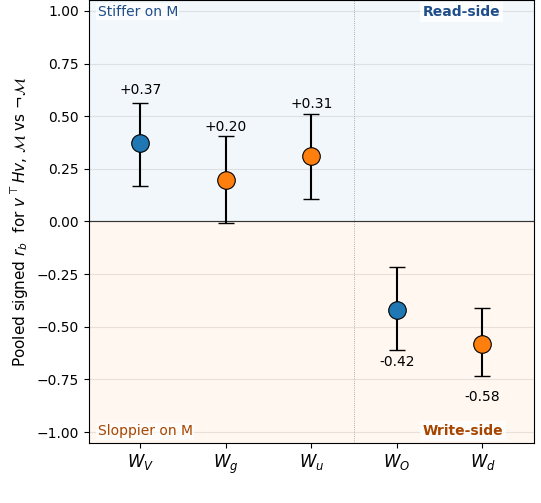}
    \caption{Directional curvature on $\Mset$ versus $c\Mset$ in the 1.28B
    base model (30 batches, 64 directions). Read-side slices are stiffer on
    $\Mset$; write-side slices are sloppier. Error bars are bootstrap $95\%$
    confidence intervals.}
    \label{fig:hessian}
    \vspace{-3.5\baselineskip}
\end{wrapfigure}

Figure~\ref{fig:hessian} shows that this geometry separates the two operator roles. Curvature is higher on $\Mset$ than on $\neg\Mset$ for the read-side matrices $W_V$, $W_{\rm gate}$, and $W_{\rm up}$. Moving these parameters is therefore more costly. The pattern reverses for the write-side matrices $W_O$ and $W_{\rm down}$, whose curvature is lower on $\Mset$. Thus, the loss constrains how the model reads MAs more strongly than how it writes to them. This geometric asymmetry matches the read-blind, write-open structure found in Section~\ref{sec:read-blindness}.

For deep strong-amplifier $\Mset$ rows of $W_{\rm down}$, the cross-entropy gradient locally favors increasing the row norm, while weight decay opposes this tendency (Appendix~\ref{sec:def:curvature} and Figure~\ref{fig:deep-write-geometry}). This partial force balance does not determine the net effect of AdamW optimizer update. 

\textbf{The full AdamW step increases massive-activation magnitude.}
We therefore measure the end effect directly. At the trained checkpoint, we apply a synthetic next AdamW step to the block that produces each target activation and compute its first-order effect on the activation's $L_2$ norm. We call this \emph{gradient pressure}; its definition and the exact AdamW decomposition are given in Appendix~\ref{sec:def:activation-pressure}. As shown in figure~\ref{fig:adamw-pressure}, the full update has positive pressure on $\Mset$ at every layer, with a larger effect in later layers, while its pressure on matched non-massive coordinates stays near zero. Decomposing the step shows that the positive pressure comes mainly from AdamW's saved moment state. The new batch gradient has a much smaller effect, and weight decay reduces rather than increases the massive activations.

Together, the two analyses answer different parts of the question. The
curvature analysis shows that massive coordinates occupy a distinct
read--write geometry in parameter space. The optimizer analysis shows the next AdamW step is poised to reinforce their activation magnitude, with the accumulated optimizer state overcoming weight decay. This is local evidence that training actively maintains massive activations rather
than merely tolerating them.

%% file: sections/8_conclusion.tex
\section{Conclusion}
Massive activations persist because the transformer architecture systematically learns to ignore them while reading, yet continues to amplify them while writing. Our findings reveal that this read-write asymmetry is not an architectural quirk, but a systems-level configuration strictly enforced by the training dynamics and loss landscape. By constraining the $W_V$, $W_{\rm up}$ and $W_{\rm gate}$ matrices, we demonstrated that MAs are not tied to a single component; rather, the network actively reorganizes read-blindness to other operators like the FFN. We speculate that read-blindness is a cheap route for the model to maintain token-independent features, which are perhaps necessary for modeling language. However, the model architecture, especially the quadratic amplifier in FFN and lack of feedback control in write operations for all operators, can amplify a subset of these features to massive values as an unintended consequence.  

Specifically, our analysis refines the directional-amplifier hypothesis. Massive coordinates do not share a distinctive amplifier direction; instead, they are distinguished by unusually high quadratic gain concentrated in their respective dominant directions. FFN read-blindness becomes detectable before this amplifier specialization, suggesting complementary rather than competing roles: the amplifier explains how selected coordinates receive exceptional FFN writes, while read-blindness explains why subsequent blocks do not condition corrective updates on the massive values already present. Within $\Mset$, stronger amplification is associated with greater concentration of the corresponding $\WD$ row, consistent with a partial FFN read pathway being concentrated into a dominant mode.

Our work provides strong evidence that the MA state is a fixed point actively maintained by the optimizer. Read-side rows exhibit high curvature, heavily penalizing deviations from the read-blind state, while write-side rows sit at a saddle point stabilized by a delicate force balance: the cross-entropy loss pulls the weights outward, while weight decay pushes them inward. AdamW optimizer actively maintains the MA via its momentum state. By identifying read-blindness as the core mechanism, we provide a compelling mechanistic explanation for why subsequent layers in a transformer fail to normalize these extreme outliers.

%% file: appendices/defs.tex
\section{Definitions and Transformer Architecture}\label{app:notation}

Let $d_m$ be the residual dimension, $H$ the number of attention heads, $d_h=d_m/H$ the dimension of each head, and $d_{ff}$ the FFN hidden dimension. We represent the residual stream after layer $l$ as $Z^l=[z^l_1,\ldots,z^l_T]\in\R^{d_m\times T}$, with one token per column. For a single Pre-LN layer with input $X$, we write
\[
Y = X + \mathrm{MHA}(\widetilde X),
\qquad
Z = Y + \mathrm{FFN}(\widetilde Y),
\]
where $X$ is the layer input and $\widetilde X=\mathrm{RMSNorm}(X)$ and $\widetilde Y=\mathrm{RMSNorm}(Y)$ are computed independently for each token. Below, we also write $Z_0=X$, $Z_1=Y$, and $Z_2=Z$.

We use column-vector linear maps of the form $y=Ax$. Each attention head $h$ therefore has read matrices $W_Q^{(h)},W_K^{(h)},W_V^{(h)}\in\R^{d_h\times d_m}$ and an output matrix $W_O^{(h)}\in\R^{d_m\times d_h}$. Defining the row-normalized attention matrix
\[
A^{(h)}(X)=\mathrm{softmax}\!\left(
\frac{(W_Q^{(h)}X)^\top(W_K^{(h)}X)}{\sqrt{d_h}}
\right),
\]
the full multi-head output is
\[
\mathrm{MHA}(X)=\sum_{h=1}^H
W_O^{(h)}W_V^{(h)}X A^{(h)}(X)^\top.
\]
We use $W_Q,W_K,W_V,W_O\in\R^{d_m\times d_m}$ to denote the corresponding stacked or concatenated MHA matrices.

The SwiGLU feed-forward block uses read matrices $W_{\text{gate}},W_{\text{up}}\in\R^{d_{ff}\times d_m}$ and an output matrix $W_{\text{down}}\in\R^{d_m\times d_{ff}}$:
\[
\mathrm{FFN}(X)=W_{\text{down}}\bigl(
\mathrm{SiLU}(W_{\text{gate}}X)\odot(W_{\text{up}}X)
\bigr).
\]

Finally, we denote the additive contributions of each module as $\Delta^{\text{attn}}=Y-X$ and $\Delta^{\text{ffn}}=Z-Y$.

\section{Gram Matrices for Read and Write Analysis}\label{app:grams}

To apply the read-blindness scores $\eta_k$ and $\mathrm{era}_k$ to a transformer, we form one Gram matrix per operator role. We use two complementary views.

\paragraph{Weight View.}
The Weight View asks what the trained weights, taken alone, permit each block to read or write. On the \emph{read side}, a feature in the null space of a weight matrix is provably invisible to that operator regardless of input. On the \emph{write side}, the Gram measures structural access to each residual output coordinate; what is actually deposited also depends on the operator's internal activations. The six Weight-View Grams, summing over heads $h$, are:
\begin{align*}
G^{a,wi} &= {\textstyle\sum_h W_V^{(h)\top}W_V^{(h)}}    & \text{($W_V$ read --- value projection)} \\
G^{a,wo} &= {\textstyle\sum_h W_O^{(h)}W_O^{(h)\top}}    & \text{($W_O$ write --- output projection)} \\
G^{f,wi} &= \WG^\top \WG + \WU^\top \WU                  & \text{(FFN gate/up read)} \\
G^{f,wo} &= \WD\,\WD^\top                                & \text{(FFN down write)} \\
G^{Q}    &= {\textstyle\sum_h W_Q^{(h)\top}W_Q^{(h)}}    & \text{($W_Q$ read --- query projection)} \\
G^{K}    &= {\textstyle\sum_h W_K^{(h)\top}W_K^{(h)}}.   & \text{($W_K$ read --- key projection)}
\end{align*}

We list $G^Q$ and $G^K$ separately from $G^{a,wi}$ because, while all three read from the residual stream, they play different computational roles: $W_V$ determines which information is routed into the attention output, whereas $W_Q$ and $W_K$ determine the attention pattern itself. Read-blindness in $W_V$ blocks information flow directly; read-blindness in $W_Q$/$W_K$ distorts where attention looks.

\paragraph{Operator View.}
The Operator View uses empirical second-moment matrices computed on a held-out corpus. Its input Grams measure which residual coordinates are available at the input to each block, while its output Grams measure which coordinates receive energy from the block's realized residual update. The four Operator-View Grams are:
\begin{align*}
G^{a,oi} &= \E\!\bigl[\tilde{Z}_0\,\tilde{Z}_0^\top\bigr]
   & \text{(Attention input second moment)} \\
G^{a,oo} &= \E\!\bigl[(Z_1-Z_0)(Z_1-Z_0)^\top\bigr]
   & \text{(Attention residual deposit)} \\
G^{f,oi} &= \E\!\bigl[\tilde{Z}_1\,\tilde{Z}_1^\top\bigr]
   & \text{(FFN input second moment)} \\
G^{f,oo} &= \E\!\bigl[(Z_2-Z_1)(Z_2-Z_1)^\top\bigr].
   & \text{(FFN residual deposit)}
\end{align*}

Expectations are taken over tokens and sequences, with each token represented as a column vector. $G^{a,oi}$ and $G^{f,oi}$ measure which features the normalized residual stream emphasizes at the input to each block; they do not by themselves measure whether the following operator reads those features. $G^{a,oo}$ and $G^{f,oo}$ measure which residual coordinates actually receive energy from each block's output.

\section{Null Occupancy}\label{app:null-occupancy}

A component of the input in the null space of $A$ has no effect on its output, so approximate null directions reveal features that the matrix effectively ignores. We use the input-side Gram matrix $G=A^\top A$, which has the same null space as $A$ and is symmetric positive semidefinite. Because the strict algebraic null space is brittle for learned matrices, we instead use an \emph{energy-thresholded} approximate null subspace.

Let $G = \sum_i \sigma_i v_i v_i^\top$ be the eigendecomposition of $G$, with eigenvalues in descending order. We define
\[
\mathcal{N}_\tau(G)
= \operatorname{span}\!\left\{\, v_i \;:\;
  \sigma_i \le \varepsilon\,\sigma_{\max}
  \ \text{ and }\
  \sum_{j \ge i} \sigma_j \le (1-\tau)\!\sum_j \sigma_j
\right\},
\]
where $\tau \in (0,1)$ controls the retained energy and $\varepsilon$ prevents spurious null detection under nearly uniform spectra. We use $\tau=0.99$ and $\varepsilon=0.01$. Let $P_{\mathcal{N}_\tau(G)}$ denote the orthogonal projector onto this subspace.

The \emph{null occupancy} of feature $k$ is the fraction of its coordinate direction $e_k$ that lies in the approximate null subspace:
\[
\eta_k(G)
= \frac{\|P_{\mathcal{N}_\tau(G)}e_k\|^2}{\|e_k\|^2}
\;\in\;[0,1].
\]
When $\eta_k(G)\approx1$, coordinate $k$ lies predominantly in directions that $A$ effectively ignores; when $\eta_k(G)\approx0$, it lies outside the approximate null subspace and is actively read.

\input{tables/detector_unified_eta_table_full.tex}

\section{Probing Results on Read-Blindness}\label{app:read-blindness-probing}
\input{tables/detector_unified_table_full_intervention.tex}
\input{tables/detector_unified_eta_table_full_intervention.tex}

%% file: tables/detector_unified_eta_table_full.tex
\begin{table}[t]
\centering
\caption{Null occupancy results. Cells report the median ${\eta}_{\mathcal{M}}$ with rank-biserial $r_b$ in parentheses. Bold cells have $r_b>0.95$; stars denote one-sided Mann--Whitney tests of ${\eta}_{\mathcal{M}}>{\eta}_{\neg\mathcal{M}}$ with $p<0.05$.}
\label{tab:read-write-null-occupancy}
\small
\setlength{\tabcolsep}{4pt}
\begin{tabular}{rcccc}
\toprule
 & Llama 135M & Llama 1.28B & Llama 2.56B & Qwen3 1.7B \\
\midrule
Massive coordinates $|\mathcal{M}|$ & 10 & 10 & 5 & 7 \\
Maximum activation magnitude & 280 & $2.27\!\times\!10^{3}$ & $3.92\!\times\!10^{3}$ & $2.23\!\times\!10^{3}$ \\
\midrule
\multicolumn{1}{l}{\emph{Input availability:}} &  &  &  &  \\
\quad Attention normalized input, $\tilde{X}$ & 0.52 (-0.40) & 0.52 (-0.24) & 0.52 (-0.56) & 0.76 (+0.19) \\
\quad FFN normalized input, $\tilde{Y}$ & 0.51 (-0.41) & \textbf{0.80 (+1.00)}\textsuperscript{*} & 0.65 (+0.93)\textsuperscript{*} & 0.80 (+0.68)\textsuperscript{*} \\
\midrule
\multicolumn{1}{l}{\emph{Read side weight view:}} &  &  &  &  \\
\quad Value projection $W_V$ & \textbf{0.69 (+0.99)}\textsuperscript{*} & \textbf{0.72 (+1.00)}\textsuperscript{*} & \textbf{0.73 (+1.00)}\textsuperscript{*} & \textbf{0.74 (+1.00)}\textsuperscript{*} \\
\quad Query projection $W_Q$ & \textbf{0.69 (+0.98)}\textsuperscript{*} & \textbf{0.70 (+0.99)}\textsuperscript{*} & \textbf{0.71 (+1.00)}\textsuperscript{*} & \textbf{0.72 (+1.00)}\textsuperscript{*} \\
\quad Key projection $W_K$ & 0.60 (+0.84)\textsuperscript{*} & \textbf{0.63 (+0.96)}\textsuperscript{*} & \textbf{0.61 (+0.98)}\textsuperscript{*} & \textbf{0.67 (+1.00)}\textsuperscript{*} \\
\quad FFN gate/up $W_{\mathrm{gate}},W_{\mathrm{up}}$ & \textbf{0.79 (+0.99)}\textsuperscript{*} & \textbf{0.87 (+1.00)}\textsuperscript{*} & \textbf{0.89 (+1.00)}\textsuperscript{*} & \textbf{0.85 (+1.00)}\textsuperscript{*} \\
\midrule
\multicolumn{1}{l}{\emph{Write side weight view:}} &  &  &  &  \\
\quad Attention output $W_O$ & 0.51 (-0.71) & 0.50 (-0.58) & 0.49 (-0.50) & 0.51 (-0.11) \\
\quad FFN down projection $W_{\mathrm{down}}$ & 0.49 (-0.84) & 0.50 (-0.26) & 0.50 (-0.49) & 0.50 (-0.16) \\
\midrule
\multicolumn{1}{l}{\emph{Write side operator view:}} &  &  &  &  \\
\quad Attention update $\Delta^{\mathrm{attn}}$ & 0.51 (-0.74) & 0.50 (-0.61) & 0.49 (-0.57) & 0.51 (-0.13) \\
\quad FFN residual update $\Delta^{\mathrm{ffn}}$ & 0.49 (-0.83) & 0.49 (-0.63) & 0.49 (-0.49) & 0.50 (-0.41) \\
\bottomrule
\end{tabular}
\end{table}

%% file: tables/detector_unified_table_full_intervention.tex
\begin{table}[t]
\centering
\caption{Llama 1.28B Erasure results for intervention settings. Bold cells have $r_b>0.95$; stars denote one-sided Mann--Whitney tests of ${\mathrm{era}}_{\mathcal{M}}>{\mathrm{era}}_{\neg\mathcal{M}}$ with $p<0.05$.}
\label{tab:read-write-erasure-intervention}
\small
\setlength{\tabcolsep}{4pt}
\begin{tabular}{rccc}
\toprule
 & $W_V$ Frozen & $W_V$ Reparam & $W_{\mathrm{up}}, W_{\mathrm{gate}}$ Frozen \\
\midrule
Massive coordinates $|\mathcal{M}|$ & 7 & 10 & 11 \\
Maximum activation magnitude & $2.08\!\times\!10^{3}$ & $2.12\!\times\!10^{3}$ & 681 \\
\midrule
\multicolumn{1}{l}{\emph{Input availability:}} &  &  &  \\
\quad Attention normalized input, $\tilde{X}$ & 0.51 (-0.52) & 0.56 (-0.57) & 0.61 (-0.17) \\
\quad FFN normalized input, $\tilde{Y}$ & 0.58 (+0.70)\textsuperscript{*} & 0.52 (-0.28) & 0.57 (-0.17) \\
\midrule
\multicolumn{1}{l}{\emph{Read side weight view:}} &  &  &  \\
\quad Value projection $W_V$ & 0.50 (+0.22) & 0.50 (+0.09) & \textbf{0.69 (+0.98)}\textsuperscript{*} \\
\quad Query projection $W_Q$ & \textbf{0.61 (+1.00)}\textsuperscript{*} & \textbf{0.64 (+0.99)}\textsuperscript{*} & \textbf{0.66 (+0.99)}\textsuperscript{*} \\
\quad Key projection $W_K$ & 0.55 (+0.42)\textsuperscript{*} & 0.55 (+0.24) & 0.63 (+0.93)\textsuperscript{*} \\
\quad FFN gate/up $W_{\mathrm{gate}},W_{\mathrm{up}}$ & \textbf{0.64 (+1.00)}\textsuperscript{*} & \textbf{0.64 (+0.99)}\textsuperscript{*} & 0.50 (-0.21) \\
\midrule
\multicolumn{1}{l}{\emph{Write side weight view:}} &  &  &  \\
\quad Attention output $W_O$ & 0.52 (-0.16) & 0.51 (-0.41) & 0.51 (-0.37) \\
\quad FFN down projection $W_{\mathrm{down}}$ & 0.49 (-1.00) & 0.50 (-0.95) & 0.50 (-0.77) \\
\midrule
\multicolumn{1}{l}{\emph{Write side operator view:}} &  &  &  \\
\quad Attention update $\Delta^{\mathrm{attn}}$ & 0.84 (-0.68) & 0.83 (-0.43) & 0.60 (-0.75) \\
\quad FFN residual update $\Delta^{\mathrm{ffn}}$ & 0.59 (-0.96) & 0.61 (-0.48) & 0.60 (-0.70) \\
\bottomrule
\end{tabular}
\end{table}

%% file: tables/detector_unified_eta_table_full_intervention.tex
\begin{table}[t]
\centering
\caption{Llama 1.28B Null occupancy results for intervention settings. Bold cells have $r_b>0.95$; stars denote one-sided Mann--Whitney tests of ${\mathrm{era}}_{\mathcal{M}}>{\mathrm{era}}_{\neg\mathcal{M}}$ with $p<0.05$.}
\label{tab:read-write-null-occupancy-intervention}
\small
\setlength{\tabcolsep}{4pt}
\begin{tabular}{rccc}
\toprule
 & $W_V$ Frozen & $W_V$ Reparam & $W_{\mathrm{up}}, W_{\mathrm{gate}}$ Frozen \\
\midrule
Massive coordinates $|\mathcal{M}|$ & 7 & 10 & 11 \\
Maximum activation magnitude & $2.08\!\times\!10^{3}$ & $2.12\!\times\!10^{3}$ & 681 \\
\midrule
\multicolumn{1}{l}{\emph{Input availability:}} &  &  &  \\
\quad Attention normalized input, $\tilde{X}$ & 0.57 (+0.30) & 0.54 (-0.13) & 0.55 (+0.02) \\
\quad FFN normalized input, $\tilde{Y}$ & \textbf{0.89 (+1.00)}\textsuperscript{*} & \textbf{0.80 (+1.00)}\textsuperscript{*} & 0.68 (+0.77)\textsuperscript{*} \\
\midrule
\multicolumn{1}{l}{\emph{Read side weight view:}} &  &  &  \\
\quad Value projection $W_V$ & 0.50 (+0.00) & 0.50 (+0.00) & \textbf{0.67 (+0.99)}\textsuperscript{*} \\
\quad Query projection $W_Q$ & \textbf{0.66 (+0.99)}\textsuperscript{*} & \textbf{0.69 (+1.00)}\textsuperscript{*} & \textbf{0.66 (+0.99)}\textsuperscript{*} \\
\quad Key projection $W_K$ & \textbf{0.59 (+0.98)}\textsuperscript{*} & \textbf{0.61 (+0.99)}\textsuperscript{*} & \textbf{0.65 (+0.97)}\textsuperscript{*} \\
\quad FFN gate/up $W_{\mathrm{gate}},W_{\mathrm{up}}$ & \textbf{0.87 (+1.00)}\textsuperscript{*} & \textbf{0.90 (+1.00)}\textsuperscript{*} & 0.50 (+0.00) \\
\midrule
\multicolumn{1}{l}{\emph{Write side weight view:}} &  &  &  \\
\quad Attention output $W_O$ & 0.52 (+0.00) & 0.51 (-0.12) & 0.51 (-0.38) \\
\quad FFN down projection $W_{\mathrm{down}}$ & 0.50 (-0.42) & 0.51 (-0.43) & 0.50 (-0.76) \\
\midrule
\multicolumn{1}{l}{\emph{Write side operator view:}} &  &  &  \\
\quad Attention update $\Delta^{\mathrm{attn}}$ & 0.52 (-0.07) & 0.52 (-0.10) & 0.51 (-0.48) \\
\quad FFN residual update $\Delta^{\mathrm{ffn}}$ & 0.49 (-0.99) & 0.50 (-0.54) & 0.49 (-0.78) \\
\bottomrule
\end{tabular}
\end{table}

%% file: appendices/implementation_details.tex
\section{Implementation and Model Details}
\label{app:implementation-details}

We follow the data and hyperparameter setup of \citet{gu2024attention} for all models.  Each model is trained on the 5-billion-token RegMix dataset for 20{,}000 optimization steps, using a context length of 2{,}048 tokens.  We use the GPT-NeoX tokenizer adopted by \citet{gu2024attention}, whose vocabulary contains 50{,}257 tokens.  Table~\ref{tab:model-architectures} summarizes the architectural configurations evaluated in our experiments. Training is done under a fixed seed.

\input{tables/model_architectures.tex}

We optimize every model with AdamW \citep{loshchilov2019adamw}, using a peak learning rate of $10^{-4}$, a weight decay of $0.01$, and momentum parameters $\beta_1=0.9$ and $\beta_2=0.999$.  The learning rate is warmed up linearly for the first 100 steps and then decayed according to a cosine schedule for the remainder of training. Models are trained on 8x8 Nvidia A100 80GB accelerator hardward. Huggingface Transformers\footnote{https://github.com/huggingface/transformers} framework is used for the model training.
To speed up the training process, liger-kernel\footnote{https://github.com/linkedin/Liger-Kernel} library is used. A typical training run concludes in ~8 hours.

%% file: tables/model_architectures.tex
\begin{table}[t]
  \caption{Llama \citep{touvron2023llama} and Qwen3 \citep{qwen3technicalreport} Architectures of the models used in our experiments.}
  \label{tab:model-architectures}
  \centering
  \small
  \setlength{\tabcolsep}{3.5pt}
  \begin{tabular}{lcccc}
    \toprule
    Property & Llama 135M & Llama 1.28B & Llama 2.56B & Qwen 1.7B \\
    \midrule
    Number of Transformer layers ($L$) & 10 & 16 & 27 & 28 \\
    Residual-stream dimension ($d_{\mathrm{model}}$) & 768 & 2{,}048 & 2{,}560 & 2{,}048 \\
    Feed-forward dimension ($d_{\mathrm{ff}}$) & 1{,}536 & 8{,}192 & 7{,}680 & 6{,}144 \\
    Number of query heads ($n_h$) & 8 & 32 & 40 & 16 \\
    Number of key--value heads ($n_{kv}$) & 8 & 32 & 40 & 16 \\
    Attention-head dimension ($d_h$) & 96 & 64 & 64 & 128 \\
    \bottomrule
  \end{tabular}
\end{table}

%% file: appendices/sstar.tex
\section{Directional FFN amplifier diagnostics}
\label{app:sstar_amplifier}

\paragraph{Directional quadratic amplifier.}
We compare our results with the directional quadratic amplifier of
\citet{sun2026massive}. Under the near-identity approximation to the SiLU
gate, let $g_i=(\WG)_{i,:}^{\top}$ and $u_i=(\WU)_{i,:}^{\top}$ denote the
input directions of intermediate unit $i$. For output coordinate $k$, define
\begin{equation}
U_k
=\sum_i \WD[k,i]g_i u_i^{\top}
=\WG^{\top}\diag\!\bigl(\WD[k,:]\bigr)\WU,
\qquad
S_k=\tfrac{1}{2}\bigl(U_k+U_k^{\top}\bigr).
\label{eq:Sk}
\end{equation}
The FFN contribution to coordinate $k$ is then approximately
\[
\mathcal{F}_{\rm ffn}(\widetilde h)_k
\approx \widetilde h^{\top}S_k\widetilde h.
\]
Let $(\lambda_\star^{(k)},s_\star^{(k)})$ be the eigenpair of $S_k$
with largest eigenvalue magnitude, with $\|s_\star^{(k)}\|_2=1$. When this
eigenpair dominates the spectrum,
\[
\mathcal{F}_{\rm ffn}(\widetilde h)_k
\approx \lambda_\star^{(k)}
\bigl(s_\star^{(k)\top}\widetilde h\bigr)^2.
\]
Thus, $s_\star^{(k)}$ is the residual-stream direction with the largest
quadratic gain into output coordinate $k$, and $\lambda_\star^{(k)}$ is its
signed gain.

\paragraph{Per-coordinate diagnostics.}
We derive four diagnostics from this construction. Except where the layer
aggregation is shown explicitly, the quantities below are computed separately
at each layer.
\begin{itemize}
    \item The overall quadratic magnitude is $\|U_k\|_F$, computed without
    explicitly assembling $U_k$ as
    \[
    \|U_k\|_F^2
    =\WD[k,:]^{\top}
    \bigl(\WG\WG^{\top}\odot\WU\WU^{\top}\bigr)\WD[k,:],
    \]
    where $\odot$ is the element-wise (Hadamard) product.

    \item The massive-subspace mass of the amplifier direction is
    \begin{equation}
    \rho_\Mset^{(k)}
    =\operatorname*{mean}_{L}
    \sum_{j\in\Mset}\bigl(s_\star^{(k,L)}[j]\bigr)^2.
    \label{eq:rho-mass}
    \end{equation}
    Because $s_\star^{(k,L)}$ is a unit vector, this is the fraction of its
    squared norm lying in the coordinate subspace
    $\operatorname{span}\{e_j:j\in\Mset\}$, averaged across layers. It is
    bounded between zero and one. An isotropically oriented unit vector has
    expected mass $|\Mset|/d_m$; larger $\rho_\Mset^{(k)}$ therefore indicates
    stronger alignment with the massive-coordinate subspace.

    In the base 1.28B model, massive output coordinates have median
    $\rho_\Mset^{(k)}=0.031$, about six times the isotropic baseline
    $10/2048\approx0.0049$. This quantity is larger than in the deterministic,
    size-matched non-massive set with rank-biserial effect $r_b=0.60$.
    Thus, their amplifier directions are preferentially aligned with the
    massive-coordinate subspace, but are not confined to it: only $3.1\%$ of
    their squared norm lies in that subspace.

    \item The leading-gain magnitude is $|\lambda_\star^{(k)}|$. Given the
    previously identified unit eigenvector, it can be computed as
    \[
    \lambda_\star^{(k)}
    =s_\star^{(k)\top}U_k s_\star^{(k)}
    =\WD[k,:]^{\top}\!\left(
      (\WG s_\star^{(k)})\odot(\WU s_\star^{(k)})
    \right).
    \]
    We report $\max_L|\lambda_\star^{(k,L)}|$ for each coordinate.

    \item The rank-one purity is
    \[
    r^{(k)}=\frac{|\lambda_\star^{(k)}|}{\|S_k\|_F}\in[0,1],
    \]
    where
    \[
    \|S_k\|_F^2
    =\tfrac{1}{2}\bigl(\|U_k\|_F^2+\tr(U_k^2)\bigr),
    \]
    and
    \[
    \tr(U_k^2)
    =\WD[k,:]^{\top}\!\bigl(
      \WG\WU^{\top}\odot\WU\WG^{\top}
    \bigr)\WD[k,:].
    \]
    Following \citet{sun2026massive}, $r^{(k)}\to1$ indicates rank-one
    dominance. The random-symmetric baseline scales as
    $\mathbb{E}[r^{(k)}]\sim2/\sqrt{d_m}$ (approximately $0.044$ for
    $d_m=2048$). We report $\max_L r^{(k,L)}$ for each coordinate.
\end{itemize}

We additionally measure direction sharing within a coordinate set
$\mathcal S$ using the sign-invariant statistic
\[
\overline{|\cos|}_{\mathcal S}
=\operatorname*{mean}_{L}
  \operatorname*{mean}_{\substack{i,j\in\mathcal S\\i\ne j}}
  \left|\left\langle
  s_\star^{(i,L)},s_\star^{(j,L)}
  \right\rangle\right|.
\]
For two independent isotropic directions in $d_m$ dimensions, the
large-$d_m$ baseline is approximately $\sqrt{2/(\pi d_m)}$.

\begin{figure}[]
    \centering
    \includegraphics[width=\linewidth]{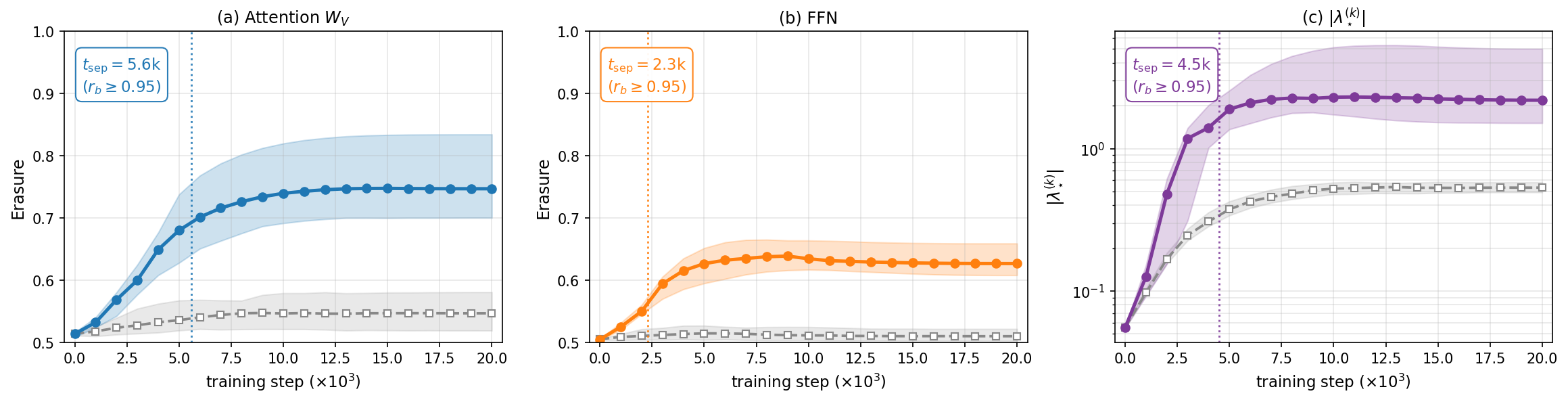}
    \caption{Emergence of $\Mset$ discrimination during pre-training of the 1.28B model (21 checkpoints, $0$–$20$k steps; per-coordinate maximum across 16 layers). Solid colored and dashed gray lines show medians and interquartile ranges for $\Mset$ ($n=10$) and sampled $c\Mset$ controls ($n=100$), respectively. The dotted line marks $t_{\rm sep}$, the first checkpoint with rank-biserial effect $>0.95$. \emph{FFN read-blindness preceeds its amplifier.}}
    \label{fig:emergence_full}
\end{figure}

\begin{figure}
    \centering
    \includegraphics[width=0.85\linewidth]{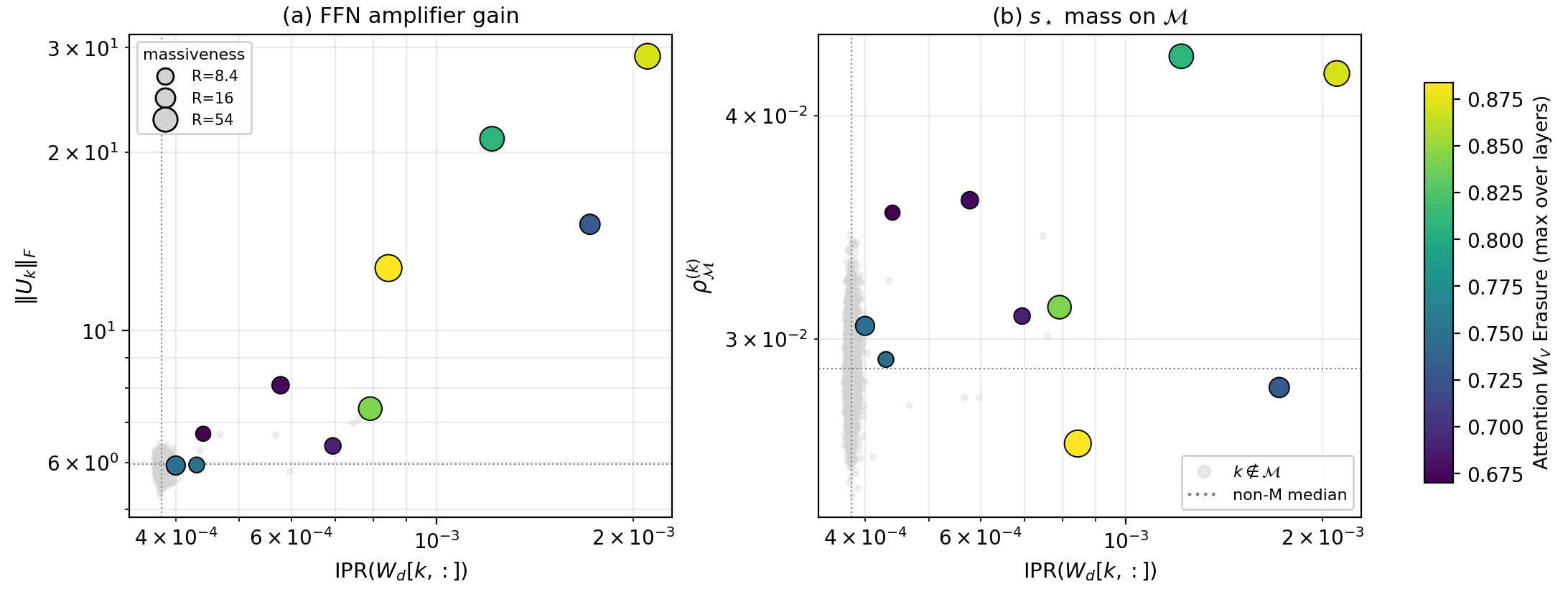}
    \caption{FFN write-row concentration vs. amplifier gain (left) and \(s_\star\) mass on \(\mathcal M\) (right). Color denotes attention-input erasure, size denotes massiveness \(R\), and grey points are non-MA coordinates.}
    \label{fig:mass_s_carries_full}
\end{figure}

%% file: appendices/gradient_analysis.tex
\section{Loss geometry and local AdamW pressure}
\label{app:gradient-analysis}

This appendix gives the definitions behind the two analyses in
Section~\ref{sec:gradient-analysis}. The curvature analysis describes the
local geometry of parameter space. The AdamW-pressure analysis instead asks
how a complete next optimizer step changes activation magnitude. These are
different measurements and we do not identify one with the other.

\subsection{Directional curvature and radial gradient}
\label{sec:def:curvature}

For each parameter slice associated with residual coordinate $k$, we sample
unit directions $v$ within that slice and measure the signed directional
curvature
\begin{equation}
    \kappa(v)=v^\top H v,
    \qquad
    H=\nabla^2\loss_{\rm CE}.
\end{equation}
We average over 64 sampled directions and compare the resulting per-coordinate
values for $\Mset$ and $\neg\Mset$ over 30 batches. Positive rank-biserial
effects mean that the signed curvature tends to be higher on $\Mset$; negative
effects mean that it tends to be lower.

The read-side matrices $W_V$, $W_{\rm gate}$, and $W_{\rm up}$ have higher
curvature on $\Mset$, whereas the write-side matrices $W_O$ and
$W_{\rm down}$ have lower curvature (Figure~\ref{fig:hessian}). For the
strong-amplifier $\Mset$ rows of $W_{\rm down}$ in deep layers, the average
signed curvature is negative. These rows therefore do not lie in an ordinary
locally convex basin along the measured directions.

\begin{figure}[t]
    \centering
    \includegraphics[width=0.65\linewidth]{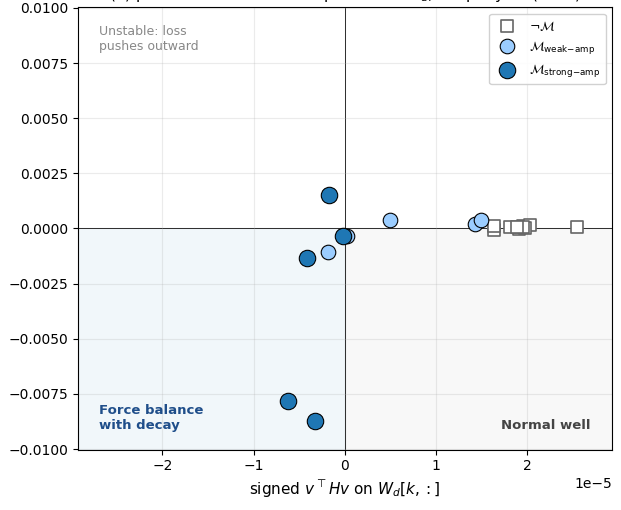}
    \caption{Curvature and radial loss-gradient projection for
    $W_{\rm down}[k,:]$ in deep layers $L\in\{9,12,15\}$. Negative
    $\nabla\loss_{\rm CE}\cdot\hat w$ means that the gradient vector points
    inward, so the loss-descent direction points outward. Strong-amplifier
    $\Mset$ coordinates concentrate in the negative-curvature,
    negative-gradient quadrant, unlike the matched non-massive coordinates.}
    \label{fig:deep-write-geometry}
\end{figure}

For the same $W_{\rm down}$ rows, we also report the radial projection
\begin{equation}
    r_k
    =\nabla_{w_k}\loss_{\rm CE}\cdot\hat w_k,
    \qquad
    \hat w_k=\frac{w_k}{\lVert w_k\rVert}.
\end{equation}
The strong-amplifier coordinates largely have $r_k<0$. This sign requires
care: the gradient vector itself points inward, but a gradient-descent update
uses the opposite direction and therefore points outward. By contrast,
AdamW's decoupled weight-decay update
$-\eta\lambda w_k$ always points inward in parameter space. The two tendencies
oppose one another. Because Figure~\ref{fig:deep-write-geometry} uses the raw loss gradient
rather than the complete preconditioned update, it establishes this local
geometry but not an exact AdamW force balance.

\subsection{AdamW update breakdown}
\label{sec:def:adamw-pressure}

We next measure the end effect of the optimizer on the activations. Let
$(m_t,v_t)$ be the saved AdamW state, let $g$ be the new local loss gradient,
and let $s=t+1$ be the next optimizer step. With
$c_1=1-\beta_1^s$ and $c_2=1-\beta_2^s$, define the adaptive update without
weight decay as
\begin{equation}
F(g;m_t,v_t)
=
-\eta
\frac{[\beta_1m_t+(1-\beta_1)g]/c_1}
{\sqrt{[\beta_2v_t+(1-\beta_2)g^2]/c_2}+\epsilon}.
\end{equation}
We decompose the next step into
\begin{align}
\Delta\theta_{\rm state}
    &=F(0;m_t,v_t),\\
\Delta\theta_{g\mid\rm state}
    &=F(g;m_t,v_t)-F(0;m_t,v_t),\\
\Delta\theta_{\rm WD}
    &=-\eta\lambda\theta,\\
\Delta\theta_{\rm full}
    &=F(g;m_t,v_t)-\eta\lambda\theta.
\end{align}
The components add exactly:
\begin{equation}
\Delta\theta_{\rm full}
=\Delta\theta_{\rm state}
+\Delta\theta_{g\mid\rm state}
+\Delta\theta_{\rm WD}.
\end{equation}
This decomposition separates the contribution already stored in the optimizer
state from the additional contribution of the current gradient and from
decoupled weight decay.

\subsection{Activation pressure}
\label{sec:def:activation-pressure}

For coordinate $k$ at the output of block $l$, define its activation magnitude
on the selected token positions $V_i$ of batch $i$ as
\begin{equation}
A_k^{(l,i)}
=
\left(\sum_{u\in V_i}h_{u,k}^{(l,i)2}\right)^{1/2}.
\end{equation}
For update component $c$, its local first-order pressure is
\begin{equation}
P_{k,c}^{(l,i)}
=
\left\langle
\nabla_{\theta^{(l)}}A_k^{(l,i)},
\Delta\theta_c^{(l,i)}
\right\rangle.
\end{equation}
Positive pressure predicts an increase in activation magnitude under that
component; negative pressure predicts a decrease. By linearity,
\begin{equation}
P_{\rm full}
=P_{\rm state}+P_{g\mid\rm state}+P_{\rm WD}.
\end{equation}
The implementation computes the Jacobian--vector product for each additive
direction once and then evaluates all selected coordinates together.

\begin{figure}[]
    \centering
    \includegraphics[width=\linewidth]{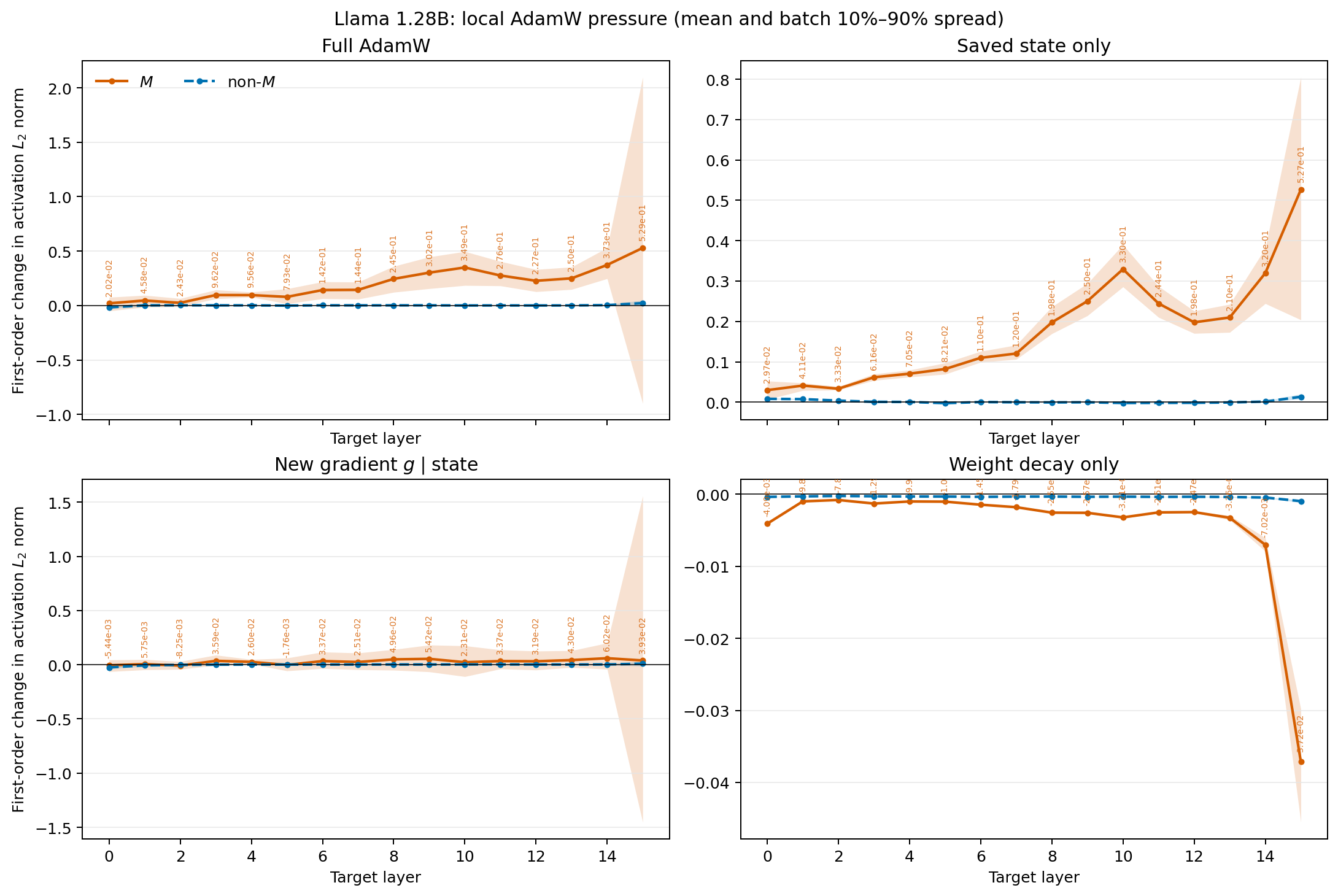}
    \caption{Local first-order AdamW pressure on activation magnitude in the
    Llama 1.28B model. Lines show the batch mean and shaded regions show the
    batch 10th--90th percentile spread. Orange denotes $\Mset$ and dashed blue
    denotes the matched non-massive control. The full AdamW step increases the
    magnitude of $\Mset$ while leaving the control near zero. This effect is
    dominated by the saved optimizer state. The current gradient conditioned
    on that state is smaller, and weight decay acts in the opposite direction.}
    \label{fig:adamw-pressure}
\end{figure}

Figure~\ref{fig:adamw-pressure} shows the layerwise result. The full AdamW step
has positive mean pressure on $\Mset$ at every layer, and the pressure generally
grows in the deeper half of the model. The matched non-massive control remains
near zero. The saved optimizer state explains most of the positive pressure;
the additional contribution of the current gradient is much smaller. Weight
decay has negative pressure on $\Mset$, especially in the final layer. Thus,
weight decay acts as a brake on massive activations, while the accumulated
adaptive state more than compensates for it.

\paragraph{Scope.}
This is a local, first-order, one-step synthetic move at a fixed checkpoint. We
update only the decoder block that produces the target activation and do not
include prefix parameters. The result therefore shows the direction in which
the next local AdamW step is poised to move the activations; it does not claim
that every step throughout training increases them.

%% file: appendices/sensitivity.tex
\section{Sensitivity to MA Selection Threshold}\label{app:sensitivity}

To evaluate sensitivity to the MA Selection threshold $R$ discussed in section~\ref{sec:methodology}, we additionally repeat erasure analysis on the Llama 1.28B model using progressively stricter thresholds. Results are reported in table~\ref{tab:r_sensitive_era}.

The main observations remain consistent across thresholds. The value, query, and FFN gate/up read projections continue to show strong positive separation between M and cM coordinates. At the more permissive threshold (R=2), the separation is slightly weaker. This is expected since M set here includes less-extreme coordinates. As the threshold increases, the read-side separation becomes stronger. Thus, the reported read-blind, write-open asymmetry is not specific to the choice and becomes more pronounced for more extreme massive activations.

\input{tables/detector_unified_R_sensitive.tex}

%% file: tables/detector_unified_R_sensitive.tex
\begin{table}[]
\centering
\caption{Erasure results at different MA selection thresholds for Llama 1.28B base model. Cells report the median ${\mathrm{era}}_{\mathcal{M}}$ with rank-biserial $r_b$ in parentheses. Bold cells have $r_b>0.95$; stars denote one-sided Mann--Whitney tests of ${\mathrm{era}}_{\mathcal{M}}>{\mathrm{era}}_{\neg\mathcal{M}}$ with $p<0.05$.}
\label{tab:r_sensitive_era}
\small
\setlength{\tabcolsep}{4pt}
\begin{tabular}{rcccc}
\toprule
 & R=2 & R=5 & R=10 & R=20 \\
\midrule
Massive coordinates $|\mathcal{M}|$ & 15 & 10 & 6 & 4 \\
Maximum activation magnitude & $2.27\!\times\!10^{3}$ & $2.27\!\times\!10^{3}$ & $2.27\!\times\!10^{3}$ & $2.27\!\times\!10^{3}$ \\
\midrule
\multicolumn{1}{l}{\emph{Input availability:}} &  &  &  &  \\
\quad Attention normalized input, $\tilde{X}$ & 0.68 (-0.66) & 0.68 (-0.73) & 0.62 (-0.74) & 0.61 (-0.77) \\
\quad FFN normalized input, $\tilde{Y}$ & 0.56 (-0.14) & 0.56 (-0.13) & 0.56 (-0.04) & 0.56 (-0.12) \\
\midrule
\multicolumn{1}{l}{\emph{Read side weight view:}} &  &  &  &  \\
\quad Value projection $W_V$ & 0.72 (+0.81)\textsuperscript{*} & \textbf{0.75 (+0.99)}\textsuperscript{*} & \textbf{0.83 (+1.00)}\textsuperscript{*} & \textbf{0.86 (+1.00)}\textsuperscript{*} \\
\quad Query projection $W_Q$ & 0.63 (+0.84)\textsuperscript{*} & \textbf{0.65 (+0.99)}\textsuperscript{*} & \textbf{0.69 (+1.00)}\textsuperscript{*} & \textbf{0.71 (+1.00)}\textsuperscript{*} \\
\quad Key projection $W_K$ & 0.54 (+0.32)\textsuperscript{*} & 0.55 (+0.31)\textsuperscript{*} & 0.53 (-0.04) & 0.52 (-0.08) \\
\quad FFN gate/up $W_{\mathrm{gate}},W_{\mathrm{up}}$ & \textbf{0.61 (+0.96)}\textsuperscript{*} & \textbf{0.63 (+1.00)}\textsuperscript{*} & \textbf{0.65 (+1.00)}\textsuperscript{*} & \textbf{0.69 (+1.00)}\textsuperscript{*} \\
\midrule
\multicolumn{1}{l}{\emph{Write side weight view:}} &  &  &  &  \\
\quad Attention output $W_O$ & 0.52 (-0.31) & 0.51 (-0.61) & 0.50 (-0.98) & 0.42 (-0.99) \\
\quad FFN down projection $W_{\mathrm{down}}$ & 0.50 (-0.57) & 0.49 (-0.64) & 0.49 (-0.96) & 0.46 (-1.00) \\
\midrule
\multicolumn{1}{l}{\emph{Write side operator view:}} &  &  &  &  \\
\quad Attention update $\Delta^{\mathrm{attn}}$ & 0.86 (-0.34) & 0.85 (-0.44) & 0.86 (-0.34) & 0.67 (-0.50) \\
\quad FFN residual update $\Delta^{\mathrm{ffn}}$ & 0.69 (-0.80) & 0.69 (-0.77) & 0.67 (-0.90) & 0.66 (-0.97) \\
\bottomrule
\end{tabular}
\end{table}